\documentclass[10pt,twocolumn,letterpaper]{article}

\usepackage[pagenumbers]{cvpr} 

\usepackage{multirow}
\usepackage{subcaption}
\usepackage{dsfont}
\usepackage{etoolbox}
\usepackage{color}

\newif\ifshowedits

\newcommand{\addeditor}[3]{%
  \definecolor{#1color}{rgb}{#3}
  \expandafter\newcommand\csname #1\endcsname[1]{%
  \ifshowedits
    {\color{#1color} ##1}%
  \else
    {##1}%
  \fi
  }%
  \expandafter\newcommand\csname #1rmk\endcsname[1]{%
  \ifshowedits
    {\color{#1color} {\bf [#2: ##1]}}
  \fi
  }%
  \expandafter\newcommand\csname #1rpl\endcsname[2]{%
  \ifshowedits
    {\color{#1color} ##1 \sout{##2}}
  \else
    {##1}
  \fi
  }%
}

\newcommand{\createtextvar}[1]{
  \expandafter\newcommand\csname #1\endcsname{%
  {\text{#1}}
}%
}
\newcommand{\textvars}[1]{\forcsvlist{\createtextvar}{#1}}

\usepackage[bigfiles]{pdfbase}
\ExplSyntaxOn
\NewDocumentCommand\embedvideo{smm}{
  \group_begin:
  \leavevmode
  \tl_if_exist:cTF{file_\file_mdfive_hash:n{#3}}{
    \tl_set_eq:Nc\video{file_\file_mdfive_hash:n{#3}}
  }{
    \IfFileExists{#3}{}{\GenericError{}{File~`#3'~not~found}{}{}}
    \pbs_pdfobj:nnn{}{fstream}{{}{#3}}
    \pbs_pdfobj:nnn{}{dict}{
      /Type/Filespec/F~(#3)/UF~(#3)
      /EF~<</F~\pbs_pdflastobj:>>
    }
    \tl_set:Nx\video{\pbs_pdflastobj:}
    \tl_gset_eq:cN{file_\file_mdfive_hash:n{#3}}\video
  }
  \pbs_pdfobj:nnn{}{dict}{
    /Type/RichMediaInstance/Subtype/Video
    /Asset~\video
    /Params~<</FlashVars (
      source=#3&
      skin=SkinOverAllNoFullNoCaption.swf&
      skinAutoHide=true&
      skinBackgroundColor=0x5F5F5F&
      skinBackgroundAlpha=0.75
    )>>
  }
  \pbs_pdfobj:nnn{}{dict}{
    /Type/RichMediaConfiguration/Subtype/Video
    /Instances~[\pbs_pdflastobj:]
  }
  \pbs_pdfobj:nnn{}{dict}{
    /Type/RichMediaContent
    /Assets~<<
      /Names~[(#3)~\video]
    >>
    /Configurations~[\pbs_pdflastobj:]
  }
  \tl_set:Nx\rmcontent{\pbs_pdflastobj:}
  \pbs_pdfobj:nnn{}{dict}{
    /Activation~<<
      /Condition/\IfBooleanTF{#1}{PV}{XA}
      /Presentation~<</Style/Embedded>>
    >>
    /Deactivation~<</Condition/PI>>
  }
  \hbox_set:Nn\l_tmpa_box{#2}
  \tl_set:Nx\l_box_wd_tl{\dim_use:N\box_wd:N\l_tmpa_box}
  \tl_set:Nx\l_box_ht_tl{\dim_use:N\box_ht:N\l_tmpa_box}
  \tl_set:Nx\l_box_dp_tl{\dim_use:N\box_dp:N\l_tmpa_box}
  \pbs_pdfxform:nnnnn{1}{1}{}{}{\l_tmpa_box}
  \pbs_pdfannot:nnnn{\l_box_wd_tl}{\l_box_ht_tl}{\l_box_dp_tl}{
    /Subtype/RichMedia
    /BS~<</W~0/S/S>>
    /Contents~(embedded~video~file:#3)
    /NM~(rma:#3)
    /AP~<</N~\pbs_pdflastxform:>>
    /RichMediaSettings~\pbs_pdflastobj:
    /RichMediaContent~\rmcontent
  }
  \phantom{#2}
  \group_end:
}
\ExplSyntaxOff

\usepackage{graphicx}

\newcommand{\mycomment}[1]{}

\newcommand{\calG}{{\cal G}}

\newcommand{\calV}{{\cal V}}

\newcommand{\btheta}{{\boldsymbol{\theta}}}

\newcommand{\bx}{{\bf x}}

\DeclareMathOperator*{\argmin}{arg\,min}

\newcommand{\vcomment}[1]{}

\addeditor{vincent}{VL}{0.0, 0.5, 0.0}
\addeditor{shiyao}{SL}{0.8, 0.4, 0.0}
\addeditor{davide}{DA}{0.6, 0.2, 0.80}

\definecolor{cvprblue}{rgb}{0.21,0.49,0.74}
\usepackage[pagebackref,breaklinks,colorlinks,allcolors=cvprblue]{hyperref}
\usepackage[accsupp]{axessibility}

\def\paperID{ } 
\def\confName{CVPR}
\def\confYear{2026}

\title{Paparazzo: Active Mapping of Moving 3D Objects}

\author{
Davide Allegro\textsuperscript{1}\quad
Shiyao Li\textsuperscript{2}\quad
Stefano Ghidoni\textsuperscript{1}\quad
Vincent Lepetit\textsuperscript{2}\\[0.9em]
\textsuperscript{1}University of Padova \\
\textsuperscript{2}LIGM, \'Ecole Nationale des Ponts et Chauss\'ees, IP Paris, Univ Gustave Eiffel, CNRS \quad
}

\begin{document}

\twocolumn[{%
\renewcommand\twocolumn[1][]{#1}%
\maketitle
\begin{center}
    \centering \captionsetup{type=figure} 
    \includegraphics[width=0.93\textwidth]{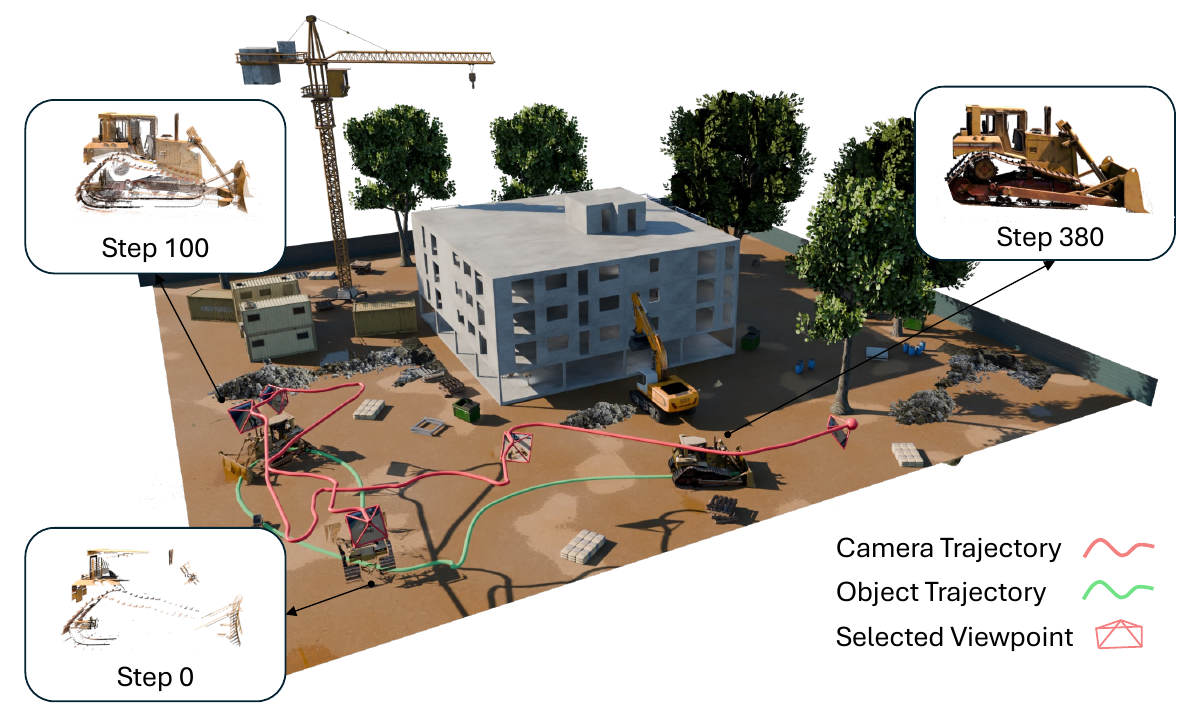} 
    \captionof{figure}{{We introduce the novel task of active mapping of moving objects, requiring agents to plan observation trajectories while compensating for target motion. We also propose a method to solve this task and a benchmark for evaluation.} }
    \label{fig:teaser}
\end{center}%
}]

\maketitle

\begin{abstract}

Current 3D mapping pipelines generally assume static environments, which limits their ability to accurately capture and reconstruct moving objects. To address this limitation, we introduce the novel task of active mapping of moving objects, in which a mapping agent must plan its trajectory while compensating for the object's motion. Our approach, Paparazzo, provides a learning-free solution that robustly predicts the target's trajectory and identifies the most informative viewpoints from which to observe it, to plan its own path. We also contribute a comprehensive benchmark designed for this new task. Through extensive experiments, we show that Paparazzo significantly improves 3D reconstruction completeness and accuracy compared to several strong baselines, marking an important step toward dynamic scene understanding. \davide{Project page: 
\url{https://davidea97.github.io/paparazzo-page/}}
\end{abstract}
    
\section{Introduction}
\label{sec:intro}

Scene exploration and mapping have been extensively studied in computer vision and robotics~\cite{yamauchi1997frontier,banta2000next,bourgault2002information}, with renewed interest driven by autonomous systems like drones and potential applications such as automated digital-twin generation. Existing exploration methods, however, rely on the assumption that the scene is static. This assumption fails in many real settings where moving objects constitute essential components of the environment. For example, in a construction site, trucks and mobile equipment are key elements of the scene that continuously reshape the workspace \davide{and accurately modeling them is important for maintaining up-to-date digital twins. However,} the site cannot be halted to capture them in static conditions. 

This is why we introduce a new task: active mapping of moving objects. An autonomous agent must reconstruct the 3D geometry of a non-cooperative object that moves independently of the mapping activity. This task is challenging because the agent must gather views that reveal new parts of the object while compensating for the object’s future motion during its own navigation. As a result, viewpoint quality depends jointly on geometric informativeness and the feasibility of reaching that view at the right time.

To solve this new task, we propose ``Paparazzo'', a learning-free framework for active 3D reconstruction of dynamic objects. Paparazzo considers a set of viewpoints  distributed in a foveal configuration around the target object and moving with it over time. To select the most informative viewpoints, we rely on Fisher Information computed from a 3D Gaussian Splatting model~\cite{jiang2024fisherrf}, while, to predict the object trajectory and the future positions of these viewpoints, we leverage an Extended Kalman Filter~\cite{ribeiro2004kalman}.
Crucially, the viewpoint with the highest expected information gain is not always the optimal choice: because viewpoints move with the target object, a view with slightly lower information gain but significantly shorter travel time may be more beneficial and efficient.
For this reason, we explore multiple strategies that jointly account for information gain and motion feasibility when selecting the next best viewpoint.
Since Paparazzo requires no training data, it generalizes to new scenes and previously unseen objects.

A first advantage of the Extended Kalman Filter~(EKF) is that it can combine past observations of the object to accurately predict its trajectory. We also use it to detect when the trajectory prediction is not reliable, e.g. when the object changes moving direction abruptly. When the EKF novelty indicates unreliable prediction, Paparazzo switches to a mode where it continuously adjusts the agent position to keep the object centered in the camera’s field of view and within an optimal range, prioritizing observations that enhance motion estimation.

For evaluation, we introduce a comprehensive benchmark and protocol for active mapping in dynamic environments, measuring reconstruction fidelity, spatial coverage, and temporal consistency across several baselines. We assume access to the target object’s mask whenever it is visible. In practice, this can be achieved by background subtraction when the static scene is known, or alternatively by using a moving-object segmentation method when it is not~\cite{tschernezki2021neuraldiff, rajivc2025segment}. For this benchmark, we developed a simulator based on the Habitat simulator~\cite{puighabitat} generating complex motions of target objects within different environments. 

Our experiments show that Paparazzo significantly improves 3D reconstruction fidelity and mapping efficiency compared to several baselines, marking a key step toward intelligent scene understanding in dynamic environments. 

Our key contributions can be summarized as:
\begin{itemize}
    \item We introduce the novel task of active mapping of moving objects, where an agent must efficiently reconstruct the 3D geometry of non-cooperative, independently moving targets.
    \item We propose Paparazzo, a learning-free dual-mode framework combining 3D Gaussian Splatting-based information gain with EKF-based motion prediction.
    \item We present the first benchmark for this task and demonstrate large performance gains across multiple dynamic scenarios.
\end{itemize}



\section{Related Work}
\label{sec:related_work}

\subsection{Active Mapping for Static Scenes}

The goal of active mapping is usually to determine how an agent should move to efficiently explore and reconstruct an unknown 3D environment. Exploration must be exhaustive: by the end of the task, the agent should have covered the entire scene while keeping its trajectory as short as possible.

Works on active mapping can be broadly categorized into traditional and learning-based approaches. Traditional methods primarily rely on heuristic strategies, such as frontier-based exploration~\cite{yamauchi1997frontier, dornhege2013frontier} and next-best-view~(NBV) selection~\cite{nbv, pito1996sensor}, or a combination of both~\cite{cao2021tare, choset2000sensor}, to guide the robot’s exploration process. They often employ voxel grids or point clouds to represent the scene.

Learning-based approaches have recently emerged to leverage deep neural networks and more expressive scene representations. For example, MACARONS~\cite{guedon2023macarons} uses neural networks to predict the coverage gain of candidate camera poses, effectively guiding NBV selection. NextBestPath~\cite{li2025nextbestpath} learns to predict the piece of trajectories that maximizes the cumulative coverage gain along the path.

With the advent of NeRF~\cite{mildenhall2021nerf} and 3D Gaussian representations~\cite{kerbl20233d}, recent works emerged, such as ANM~\cite{yan2023active}, NARUTO~\cite{feng2024naruto}, and ActiveGS~\cite{jin2025activegs}, that train such models as the intermediate scene state, using measures such as confidence or Fisher information to determine the next-best pose. Combined with traditional path planning algorithms, these methods achieve impressive performance in producing high-quality 3D reconstructions.

To the best of our knowledge, all active mapping works consider a static scene. In this paper, we are interested in mapping a non-cooperative mobile object, which is much more challenging as we need to estimate the object motion and compensate for it when planning the next move of our agent.

\subsection{Mobile Object Passive Reconstruction}

Our work is related to the reconstruction of mobile objects, such as in-hand scanning, where a target object is moved in front of one or several cameras~\cite{rusinkiewicz-02-realtime3dmodelacquisition,Weise,Wang2021DemoGraspFL,tzionas-iccv15-3dobjectreconstruction,hampali-cvpr23-inhand,jin-iccv25-6dopegs}. Like us, they aim to reconstruct an object while estimating its motion within a scene. In particular, \cite{jin-iccv25-6dopegs} also uses Gaussian primitives to represent the object as we do. The key difference is that in the case of in-hand scanning, a user moves the object aiming to improve the reconstruction, which means the object motion is intended to support the task, so it is ``cooperative''. In our case, we need to plan how to move the agent in the environment to capture new relative poses between the object and the agent, in addition to track and reconstruct the object. In such a scenario, the object moves in a ``non-cooperative'' manner, meaning that the object does not move in a way that facilitates its reconstruction.


\textvars{EIG,eig,tr,meas,cand,train}
\textvars{sync,agent,obj}

\section{Paparazzo}
\label{sec:method}

\begin{figure*}
    \centering
    \includegraphics[width=1.0\linewidth]{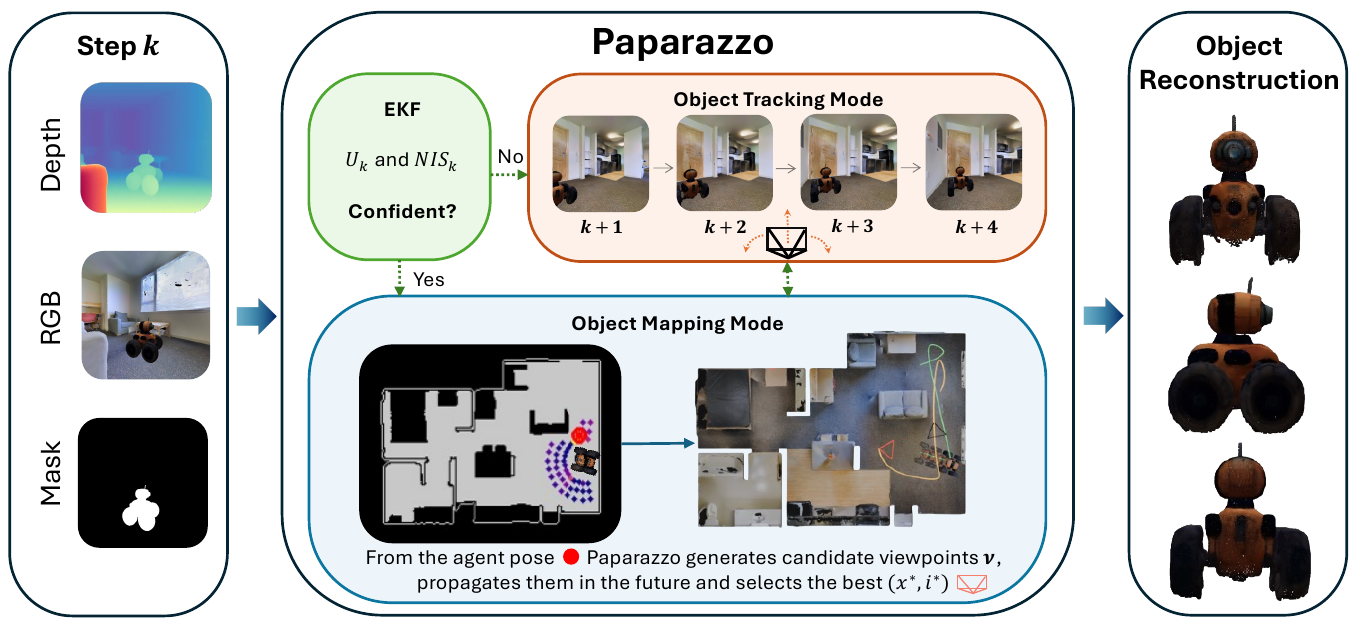}
    \caption{
    \label{fig:pipeline}
    Paparazzo alternates between Object Tracking Mode and Object Mapping Mode based on the confidence of the EKF motion estimate. When the filter is uncertain, the agent prioritizes acquiring stabilizing observations; once confident, it predicts future object motion, generates and propagates candidate viewpoints, and selects the optimal one ($\mathbf{x}^*, i^*$) that minimizes the final cost function.
    }
\end{figure*}

As shown in Figure~\ref{fig:pipeline}, our ``Paparazzo'' method alternates  between two operating phases depending on the confidence of its estimate of the target object's motion:
\begin{itemize}
    \item \emph{Object Tracking Mode}~(Section~\ref{sec:track_mode}): 
    Paparazzo switches to this mode when its estimate of the object's motion is uncertain. It then keeps the target object in its field of view to improve this estimate.
    \item \emph{Object Mapping Mode}~(Section~\ref{sec:map_mode}): Paparazzo switches to this mode when it is confident enough of the object's motion. It then plans motions to informative viewpoints for efficient reconstruction and executes them.
\end{itemize}


\subsection{Problem Formulation}
Let an agent equipped with a fixed, front-facing RGB-D camera $C$ operate in a 3D world frame $W$ containing a dynamic object $O$. 
\davide{We denote $T_A^B \in SE(3)$ as the rigid transformation from frame $A$ to frame $B$, represented as a $4\times4$ homogeneous matrix.}
At each discrete time step $k$, the camera pose $T_{C_k}^W$ is assumed to be known from the agent’s localization system, while the object pose $T_{O_k}^W$ is unknown and must be estimated.
When the object is detected, its segmentation mask $\mathcal{M}_k$ allows extracting the corresponding 3D points $\mathcal{P}_{O_{k}}^{C_{k}}=\{p^{C_{k}}_j \}_{j=1}^{P}\,$ in the camera frame. Each 3D point $p_j^{C_k}\in\mathbb{R}^{3}$ is  obtained by back-projecting the pixels within $\mathcal{M}_k$ using the available depth information and the camera intrinsic parameters.


The objective is to determine informative viewpoints that can be reached by the agent and enable it to efficiently observe and reconstruct the complete surface of the moving object with minimal views, producing a consistent 3D model in the object's local reference frame while predicting and adapting to its motion.

\subsection{Initialization}
\label{sec:initialization}
At the first detection time $t_d$, we initialize the object pose from the 3D object points $\mathcal{P}_{O_{t_{d}}}^{C_{t_{d}}}\,$. The object translation $t_{O_{t_{d}}}^{C_{t_{d}}}\in\mathbb{R}^{3}$ is defined as the centroid of these points. The object rotation $R_{O_{t_d}}^{C_{t_d}}\in SO(3)$ is constructed by aligning its $z$-axis with the world vertical direction, while the $x$–$y$ axes are obtained by performing PCA on the points projected onto the ground plane. 
The initial object pose expressed in the world coordinate system therefore is:
\begin{equation}
    T_{O_{t_d}}^W = T_{C_{t_d}}^W T_{O_{t_{d}}}^{C_{t_d}}\,,\quad\text{with}\quad T_{O_{t_d}}^{C_{t_d}} = 
    \begin{bmatrix} R_{O_{t_d}}^{C_{t_d}} & t_{O_{t_{d}}}^{C_{t_{d}}} \\ 0 & 1 \end{bmatrix}.
    \label{eq:initialization}
\end{equation}
We initialize Gaussian primitives $\mathcal{G}_O$ from the object’s segmented RGB-D observation, as in the SplaTAM backbone~\cite{keetha2024splatam}. Although SplaTAM was originally designed for static scenes, we expressed $\mathcal{G}_O$ in the object reference frame by means of the estimated object pose that remains consistent across time as the object moves.

\subsection{EKF-Based Motion Prediction}
\label{sec:ekf}

We rely on an Extended Kalman Filter (EKF) defined on $SE(3)$ to estimate the object state, composed of the object pose $T_{O_k}^W$ and its linear and angular velocities, together with its associated covariance matrix $P_k$.

We quantify our confidence in the estimated object state with two complementary metrics. The first is $U_k = \mathrm{tr}(P_k)$, which provides a compact measure of the state uncertainty. The second is the Normalized Innovation Squared (NIS), which quantifies the consistency of a new measurement $T_{O_k}^{W,\text{meas}}$ of the target object pose with the current predicted object pose $T_{O_{k|k-1}}^{W}$:
\begin{equation}
    \mathrm{NIS}_k = y_k^\top S_k^{-1} y_k\,,
\end{equation}
where $y_k = \log((T_{O_{k|k-1}}^W)^{-1} T_{O_k}^{W,\text{meas}})^\vee$
is the innovation on $SE(3)$ and $S_k = H P_{k|k-1} H^\top + R$ is the corresponding innovation covariance.

We consider the EKF estimate reliable when $U_k < \tau_u$ and $\mathrm{NIS}_k < \tau_n$ for $N_s$ consecutive steps. If this condition is met, Paparazzo switches to \emph{Object Mapping Mode} to perform information-driven active exploration and refine the reconstruction. 
Otherwise, the system transitions to the \emph{Object Tracking Mode} to re-localize the object and stabilize the EKF. 

\subsection{Object Tracking Mode}
\label{sec:track_mode}

The goal of this mode is to prioritize frequent observations of the target object in order to refine motion estimates. To this end, the agent actively keeps the object within the camera’s field of view while continuously updating its reconstruction and motion estimate. At each time step, the agent rotates to move the segmentation mask toward the image center, and translates to adjust its distance to the object so that the object's apparent size remains approximately half of the image.

We also estimate the object pose $T_{O_k}^{W,\text{meas}}$. This is done by aligning the segmented point cloud $\mathcal{P}_{O_{k}}^{C_{k}}$ 
with the object reconstruction accumulated up to time $k-1$. To this end, we first use KISS-Matcher~\cite{lim2025kiss} to obtain a coarse but globally consistent registration, robust to outliers and large displacement, and then refining this transformation using Colored ICP~\cite{korn2014color}. 

We then use $T_{O_k}^{W,\text{meas}}$ to update the EKF and improve the estimate of the object state, and to integrate the newly observed object point cloud into the object reconstruction.
The Gaussian Splatting model $\mathcal{G}_O$ is concurrently refined using the SplaTAM optimization process, which incrementally densifies and updates the dynamic object representation using the new RGB-D observation transformed into the object frame.

Once the state uncertainty stabilizes, Paparazzo shifts its focus from re-localization to information-driven exploration, leveraging the learned object GS model to guide active reconstruction.

\subsection{Object Mapping Mode}
\label{sec:map_mode}

When the EKF stabilizes, Paparazzo transitions to the \emph{Object Mapping Mode}. The goal of this mode is to move the agent to poses that will significantly improve the object reconstruction, while taking into account the object motion predicted by the EKF.

As shown in Figure~\ref{fig:pipeline}, we sample candidate viewpoints $\mathcal{V}$ relative to the object reference frame, so that they move together with it. The camera centers corresponding to these viewpoints are distributed around the object in a foveated configuration, and the cameras point toward the object.

If the object were static, we could simply select the most informative viewpoints in $\cal{V}$ according to FisherRF~\cite{jiang2024fisherrf} applied to the object GS model $\mathcal{G}_O$. However, since the object is moving, we must trade off between (i) the informativeness of a viewpoint and (ii) the temporal synchronization between the agent and the moving object. To quantify this trade-off, we introduce the following criterion:
\begin{equation}
B(\bx,i) = - w_\eig \EIG(\bx) + w_\sync C_\sync(\bx, i)  \, ,
\label{eq:cost_function_fin}
\end{equation}
%
where $\EIG(\bx)$ is the FisherRF informativeness associated with the candidate viewpoint $\bx\in \mathcal{V}$,
and $C_\sync(\bx, i)$ is a criterion we introduce to measure how well the agent can synchronize with the motion predicted for the object when attempting to observe the object from viewpoint $\bx$.
Weights $w_\sync$ and $w_\eig$ balance the contribution of the two terms.  

The FisherRF criterion quantifies how much a new viewpoint contributes to refining the parameters $\btheta$ of the current Gaussian Splatting representation $\calG_O$ of the object. It can be computed analytically and efficiently from $\calG_O$. More details can be found in \cite{jiang2024fisherrf}.

The term $C_\sync(\bx, i)$ is defined as:
\begin{equation}
    C_\sync(\bx,i) = 
    \big| \hat{s}_\agent(\bx, i) - (i - k) \big|\,.
\end{equation}
Here, $\hat{s}_\agent(\bx, i)$ denotes the number of motion steps required for the agent to reach the camera pose $T_{O_i}^W \cdot \bx$, where $T_{O_i}^W$ is the object pose predicted by the EKF for future time step $i$. We compute $\hat{s}_\agent(\bx,i)$ using an A* motion planner~\cite{ju2020path}. The term $i-k$ denotes instead the number of predicted time steps required for the object to evolve from its current pose $T_{O_k}^W$ to the predicted pose $T_{O_i}^W$. Thus, \(C_\sync(\bx,i)\) measures the temporal mismatch between the agent reaching the viewpoint associated with the object at time step $i$ and the object itself reaching its predicted pose at that same time.
We finally select the viewpoint that yields the best trade-off over a horizon of $N_h$ future time steps:
\begin{equation}
    (\bx^*, i^*) = 
    \argmin_{\substack{\bx\in\calV, (i-k)\leq N_h}} B(\bx,i) \, .
\end{equation}

While moving the agent to pose $T^W_{O_i^*}\cdot\bx^*$, Paparazzo continuously integrates new RGB-D observations into the object point cloud $\mathcal{P}$, updates $\mathcal{G}_O$, and monitors the EKF consistency.
If the NIS or state uncertainty exceeds their confidence thresholds, mapping is halted, and the system reverts to \emph{Object Tracking Mode}, maintaining a reactive loop between mapping and tracking. This dynamic coupling of information-driven mapping and motion-aware prediction constitutes the core novelty of Paparazzo, enabling 3D reconstruction of moving objects without assuming static scenes. Notably, Paparazzo runs online at 8 FPS. Full runtime and memory details are reported in the supplementary material.

\section{Experimental Results}
\label{sec:exp_and_res}

To evaluate our Paparazzo method, we introduce a dedicated benchmark and evaluation protocol designed to assess both reconstruction fidelity and spatial coverage over time. Experiments are conducted within Habitat 3.0~\cite{puighabitat}, a high-performance 3D simulator that provides realistic indoor environments and robot displacements.

\begin{figure}[t]
    \centering
    \begin{subfigure}[b]{0.48\linewidth}
        \centering
        \includegraphics[width=\linewidth]{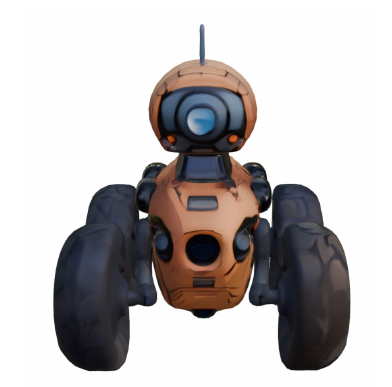}
        \caption{Object 1}
    \end{subfigure}
    \begin{subfigure}[b]{0.48\linewidth}
        \centering
        \includegraphics[width=\linewidth]{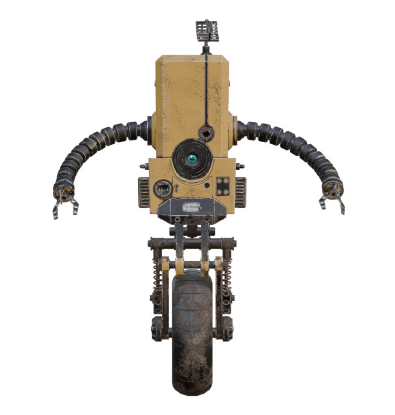}
        \caption{Object 2}
    \end{subfigure}

    \vspace{1mm}

    \begin{subfigure}[b]{0.48\linewidth}
        \centering
        \includegraphics[width=\linewidth]{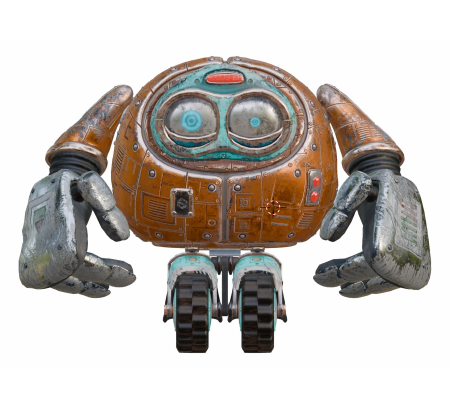}
        \caption{Object 3}
    \end{subfigure}
    \begin{subfigure}[b]{0.48\linewidth}
        \centering
        \includegraphics[width=\linewidth]{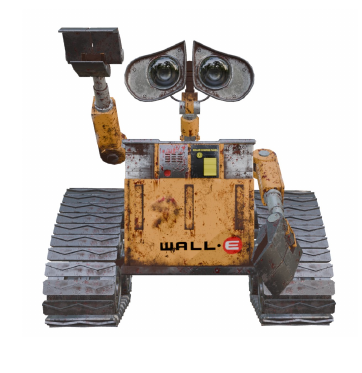}
        \caption{Object 4}
    \end{subfigure}

    \caption{The four target objects used in our experiments, featuring different shapes and colors. Each object is evaluated independently across all environments.}
    \label{fig:benchmark_objects}
\end{figure}

We selected six photorealistic indoor scenes---three from the Matterport3D dataset~(M)~\cite{chang2017matterport3d} and three from the Gibson dataset~(G)~\cite{xia2018gibson}---commonly used for static active mapping~\cite{yan2023active}.
To extend these static scenes to dynamic scenarios, we introduce a synthetic moving target object into each environment.
We consider the four target objects shown in Figure~\ref{fig:benchmark_objects}. Each object is evaluated independently across all scenes in separate runs. This setup ensures statistical diversity across both objects and environments.
The agent is equipped with an RGB-D sensor and initialized in a navigable pose, with the target object placed in front of it in a random position and orientation.

\paragraph{Object Motion Protocol.}
To comprehensively assess reconstruction performance under diverse object motion dynamics, we consider four  motion patterns for the target:
\begin{itemize}
\item \textbf{Bouncing Ball:} upon collision, the object randomly changes orientation and continues in the new direction.
\item \textbf{Forward \& Backward:} the object moves along a straight line without changing orientation, moving forward until collision and then reversing direction.

\item \textbf{Stop \& Go:} similar to \textit{Bouncing Ball}, but with intermittent stops---pausing every $S$ steps and resuming after $G$ steps---to simulate non-uniform velocity.

\davide{\item \textbf{Curved Bouncing Ball:} the object follows curved trajectories; upon collision, it randomly changes orientation and speed, then restarts along a new curved direction.}
\end{itemize}
Each motion pattern introduces distinct challenges related to motion predictability, visibility continuity, and temporal occlusions. 
Each translation of the object is 5\,cm per step, each rotation is $10^\circ$ per frame.

\paragraph{Agent Motion Model.}
The agent navigates using a discrete action space made of standard motion primitives:
\begin{itemize}
\item \textbf{Forward:} move ahead by 15\,cm,
\item \textbf{Rotate Left / Right:} yaw rotation by $10^\circ$.
\end{itemize}
These control primitives follow the canonical Habitat discretization for active perception tasks, allowing reproducible and physically plausible exploration behavior.

\subsection{Evaluation Protocol}

We report both geometric accuracy and exploration efficiency across all experiments.
Metrics are computed by comparing the reconstructed point cloud $\mathcal{P}$ against the ground-truth object model $\mathcal{P}_{GT}$ at the final time step and throughout exploration. We considered three metrics:
\begin{itemize}
\item \textbf{3D Coverage~[\%]}: the percentage of ground-truth points that have at least one reconstructed point within a distance threshold $\tau=1\,\text{cm}$~(higher = more complete reconstruction);
\item \textbf{Completeness~[cm]}: the mean distance from each ground-truth point to its nearest reconstructed point~(lower = fewer missing regions).
\item \textbf{AUC}: the area-under-curve of coverage with respect to the agent steps, reflecting how efficiently coverage increases during exploration~(higher = faster coverage).
\end{itemize}
All metrics are tested across all the scenes, objects, and motion patterns detailed above.

\paragraph{Baselines.}
We compare Paparazzo against three baselines designed to isolate the contributions of viewpoint selection, motion prediction, and temporal feasibility:
\begin{itemize}
\item Random Walk~(RW): a classical baseline in active mapping for static scenes.
The agent moves randomly across the environment, accumulating object point clouds whenever the object falls within its field of view, without considering the object motion.
\item Random Informative Selection~(RIS): an ablation of our method that selects, at each mapping iteration, a random feasible pose among the $N_h \times |\mathcal{V}|$ informative candidate viewpoints, ignoring both the synchronization cost and the predicted feasibility of observing the object from that position.
\item Tracking-Only~(TO): we keep the agent in Object Tracking Mode, a purely passive strategy that continuously tracks the object’s motion using the EKF but performs no active viewpoint selection or mapping, serving as a lower bound for reconstruction completeness.
\end{itemize}
These baselines allow us to disentangle the impact of Paparazzo’s key components---motion-aware viewpoint selection and temporal feasibility reasoning---on overall reconstruction quality and mapping efficiency.

\subsection{Results}

\vincent{
Table~\ref{tab:motion_scenes} gives the quantitative results across six different indoor scenes for our four dynamic motion types. All reported values are averaged over all test objects within each scene and across five  runs, with each run consisting of 500 agent steps.
}
For all experiments, we configured Paparazzo to predict up to $N_{h}=60$ future steps using the EKF.

Over all configurations, Paparazzo consistently outperforms the baselines in terms of coverage, completeness, and AUC, demonstrating its superior efficiency in reconstructing dynamically moving objects. Its adaptive strategy, alternating between object tracking and mapping modes, allows it to handle different motion complexities more effectively than static or passive baselines.

\begin{table*}[t!]
  \centering
  \caption{
  Quantitative results across scenes for the four dynamic motion types (Bouncing Ball~(BB), Curved Bouncing Ball~(CBB), Forward \& Backward~(FB), and Stop \& Go~(SG)). Reported values are averaged over all test objects and runs. Each entry shows Coverage (\%), Completeness (cm), and AUC.
  \label{tab:motion_scenes}
  }
  \resizebox{\textwidth}{!}{
  \begin{tabular}{ll|ccc|ccc|ccc|ccc|ccc|ccc|ccc}
    \toprule
    \multicolumn{2}{c|}{\textbf{Motion / Method}} & \multicolumn{3}{c|}{\textbf{Denmark (G)}} & \multicolumn{3}{c|}{\textbf{Ribera (G)}} & \multicolumn{3}{c|}{\textbf{Greigsville (G)}} & \multicolumn{3}{c|}{\textbf{PuKPg4mmafe (M)}} & \multicolumn{3}{c|}{\textbf{GdvgFV5R1Z5 (M)}} & \multicolumn{3}{c|}{\textbf{pLe4wQe7qrG (M)}} & \multicolumn{3}{c}{\textbf{Average}}\\
    \cmidrule(lr){3-5} \cmidrule(lr){6-8} \cmidrule(lr){9-11} \cmidrule(lr){12-14} \cmidrule(lr){15-17} \cmidrule(lr){18-20} \cmidrule(lr){21-23}
     &  & Cov & Comp & AUC & Cov & Comp & AUC & Cov & Comp & AUC & Cov & Comp & AUC & Cov & Comp & AUC & Cov & Comp & AUC & Cov & Comp & AUC\\
    \midrule
    \multirow{4}{*}{\rotatebox{90}{BB}} & RW & 57.21 & 1.79 & 0.56 & 55.93 & 1.81 & 0.55 & 49.01 & 2.20 & 0.48 & 51.10 & 2.16 & 0.50 & 51.46 & 2.02 & 0.50 & 44.31 & 2.49 & 0.44 & 51.50 & 2.08 & 0.51\\
     & RIS & 74.32 & 0.80 & 0.65 & 64.37 & 1.27 & 0.58 & 65.60 & 1.13 & 0.61 & 63.73 & 1.24 & 0.59 & 70.83 & 0.96 & 0.61 & 63.55 & 1.32 & 0.57 & 67.07 & 1.12 & 0.60\\
     & TO & 83.08 & 0.66 & 0.77 & 76.10 & 0.87 & 0.69 & 71.20 & 1.06 & 0.66 & 75.71 & 0.89 & 0.68 & 75.85 & 0.90 & 0.71 & 73.41 & 1.03 & 0.69 & 75.89 & 0.90 & 0.70\\
     & Paparazzo & \textbf{86.93} & \textbf{0.61} & \textbf{0.81} & \textbf{80.28} & \textbf{0.81} & \textbf{0.73} & \textbf{77.13} & \textbf{0.88} & \textbf{0.72} & \textbf{79.30} & \textbf{0.83} & \textbf{0.73} & \textbf{83.10} & \textbf{0.76} & \textbf{0.77} & \textbf{82.31} & \textbf{0.71} & \textbf{0.74} & \textbf{81.51} & \textbf{0.77} & \textbf{0.75}\\
    \midrule
    \multirow{4}{*}{\rotatebox{90}{CBB}} & RW & 49.32 & 2.19 & 0.49 & 58.17 & 1.76 & 0.57 & 48.48 & 2.31 & 0.48 & 51.55 & 2.18 & 0.51 & 52.85 & 2.00 & 0.52 & 47.14 & 2.27 & 0.47 & 51.25 & 2.12 & 0.50\\
     & RIS & 67.69 & 1.00 & 0.63 & 62.24 & 1.28 & 0.55 & 56.40 & 1.52 & 0.54 & 65.06 & 1.13 & 0.60 & 58.60 & 1.47 & 0.56 & 53.34 & 1.72 & 0.51 & 60.56 & 1.35 & 0.56\\
     & TO & 70.46 & 1.01 & 0.67 & \textbf{70.84} & \textbf{1.05} & 0.62 & 68.88 & 1.05 & 0.66 & 68.62 & 1.19 & 0.64 & 68.76 & \textbf{1.04} & 0.65 & 61.33 & 1.32 & 0.59 & 68.15 & \textbf{1.11} & 0.64\\
     & Paparazzo & \textbf{74.99} & \textbf{0.96} & \textbf{0.71} & 67.52 & 1.28 & \textbf{0.63} & \textbf{72.30} & \textbf{1.01} & \textbf{0.70} & \textbf{72.39} & \textbf{1.08} & \textbf{0.68} & \textbf{71.48} & 1.05 & \textbf{0.68} & \textbf{65.66} & \textbf{1.28} & \textbf{0.62} & \textbf{70.73} & \textbf{1.11} & \textbf{0.67}\\
    \midrule
    \multirow{4}{*}{\rotatebox{90}{FB}} & RW & 50.03 & 2.16 & 0.49 & 53.21 & 1.91 & 0.53 & 47.44 & 2.35 & 0.47 & 53.25 & 1.98 & 0.52 & 51.49 & 2.02 & 0.51 & 46.01 & 2.35 & 0.46 & 50.24 & 2.13 & 0.50\\
     & RIS & 48.06 & 2.17 & 0.46 & 54.61 & 1.70 & 0.49 & 59.65 & 1.51 & 0.53 & 58.49 & 1.56 & 0.53 & 52.07 & 1.80 & 0.49 & 54.85 & 1.73 & 0.51 & 54.62 & 1.75 & 0.50\\
     & TO & 53.34 & 1.93 & 0.52 & 66.45 & 1.23 & 0.61 & 59.44 & 1.49 & 0.56 & 66.27 & 1.22 & 0.60 & 64.36 & 1.35 & 0.59 & 62.44 & 1.47 & 0.58 & 62.05 & 1.45 & 0.58\\
     & Paparazzo & \textbf{60.83} & \textbf{1.53} & \textbf{0.57} & \textbf{71.13} & \textbf{1.11} & \textbf{0.63} & \textbf{72.45} & \textbf{1.04} & \textbf{0.66} & \textbf{67.01} & \textbf{1.20} & \textbf{0.63} & \textbf{69.60} & \textbf{1.18} & \textbf{0.62} & \textbf{65.34} & \textbf{1.35} & \textbf{0.61} & \textbf{67.73} & \textbf{1.23} & \textbf{0.62}\\
    \midrule
    \multirow{4}{*}{\rotatebox{90}{SG}} & RW & 56.59 & 1.82 & 0.56 & 56.53 & 1.83 & 0.55 & 49.09 & 2.24 & 0.48 & 52.42 & 2.13 & 0.51 & 51.90 & 1.98 & 0.51 & 44.28 & 2.48 & 0.44 & 51.80 & 2.08 & 0.51\\
     & RIS & 56.98 & 1.58 & 0.56 & 44.68 & 2.33 & 0.44 & 49.78 & 2.16 & 0.49 & 45.61 & 2.33 & 0.45 & 48.09 & 2.10 & 0.47 & 46.31 & 2.33 & 0.46 & 48.58 & 2.14 & 0.48\\
     & TO & 68.94 & 1.05 & 0.67 & 58.16 & 1.54 & 0.56 & 60.90 & 1.42 & 0.60 & 55.47 & 1.71 & 0.54 & 63.97 & 1.25 & 0.62 & 57.29 & 1.51 & 0.56 & 60.79 & 1.41 & 0.59\\
     & Paparazzo & \textbf{79.70} & \textbf{0.79} & \textbf{0.77} & \textbf{71.20} & \textbf{1.12} & \textbf{0.65} & \textbf{62.85} & \textbf{1.38} & \textbf{0.62} & \textbf{56.03} & \textbf{1.69} & \textbf{0.55} & \textbf{72.22} & \textbf{1.04} & \textbf{0.68} & \textbf{66.19} & \textbf{1.20} & \textbf{0.63} & \textbf{68.03} & \textbf{1.20} & \textbf{0.65}\\
    \bottomrule
  \end{tabular}
  }
\end{table*}

\begin{table*}[t]
  \centering
  \caption{
  Quantitative comparison of different motion types averaged across all scenes. We evaluate four dynamic behaviors (Bouncing Ball~(BB), Curved Bouncing Ball~(CBB), Forward \& Backward~(FB), and Stop \& Go~(SG)) and report results for all test objects. Each cell shows Coverage (\%), Completeness (cm), and AUC.
  \label{tab:motion_objects}
  }
  \resizebox{\textwidth}{!}{
  \begin{tabular}{@{}ll|ccc|ccc|ccc|ccc|ccc@{}}
    \toprule
    \multicolumn{2}{c|}{\textbf{Motion / Method}} & \multicolumn{3}{c|}{\textbf{Object 1}} & \multicolumn{3}{c|}{\textbf{Object 2}} & \multicolumn{3}{c|}{\textbf{Object 3}} & \multicolumn{3}{c|}{\textbf{Object 4}} & \multicolumn{3}{c}{\textbf{Average}} \\
    \cmidrule(lr){3-5} \cmidrule(lr){6-8} \cmidrule(lr){9-11} \cmidrule(lr){12-14} \cmidrule(lr){15-17}
     &  & Cov & Comp & AUC & Cov & Comp & AUC & Cov & Comp & AUC & Cov & Comp & AUC & Cov & Comp & AUC \\
    \midrule
    \multirow{4}{*}{\rotatebox{90}{BB}} & RW & 57.30 & 1.84 & 0.56 & 64.19 & 1.24 & 0.64 & 43.99 & 2.64 & 0.43 & 40.54 & 2.59 & 0.40 & 51.50 & 2.08 & 0.51\\
     & RIS & 75.84 & 0.78 & 0.68 & 63.74 & 1.14 & 0.60 & 68.24 & 1.10 & 0.59 & 60.45 & 1.45 & 0.53 & 67.07 & 1.12 & 0.60\\
     & TO & 86.94 & 0.60 & 0.80 & 71.89 & 0.95 & 0.69 & 78.25 & 0.86 & 0.71 & 66.49 & 1.19 & 0.59 & 75.89 & 0.90 & 0.70\\
     & Paparazzo & \textbf{90.64} & \textbf{0.55} & \textbf{0.83} & \textbf{79.19} & \textbf{0.78} & \textbf{0.75} & \textbf{82.60} & \textbf{0.73} & \textbf{0.75} & \textbf{73.60} & \textbf{1.00} & \textbf{0.67} & \textbf{81.51} & \textbf{0.77} & \textbf{0.75}\\
    \midrule
    \multirow{4}{*}{\rotatebox{90}{CBB}} & RW & 56.04 & 2.07 & 0.55 & 62.73 & 1.33 & 0.62 & 43.68 & 2.59 & 0.43 & 42.55 & 2.49 & 0.42 & 51.25 & 2.12 & 0.50\\
     & RIS & 66.90 & 1.07 & 0.62 & 62.87 & 1.18 & 0.61 & 58.26 & 1.61 & 0.53 & 54.20 & 1.55 & 0.51 & 60.56 & 1.35 & 0.56\\
     & TO & 76.13 & 0.87 & 0.71 & 67.98 & 1.09 & 0.66 & 67.36 & \textbf{1.13} & 0.61 & 61.12 & 1.34 & 0.57 & 68.15 & \textbf{1.11} & 0.64\\
     & Paparazzo & \textbf{79.13} & \textbf{0.83} & \textbf{0.74} & \textbf{71.39} & \textbf{1.04} & \textbf{0.69} & \textbf{68.70} & 1.24 & \textbf{0.64} & \textbf{63.68} & \textbf{1.33} & \textbf{0.60} & \textbf{70.72} & \textbf{1.11} & \textbf{0.67}\\
    \midrule
    \multirow{4}{*}{\rotatebox{90}{FB}} & RW & 53.81 & 1.87 & 0.53 & 65.02 & 1.22 & 0.64 & 41.28 & 2.84 & 0.41 & 40.84 & 2.58 & 0.41 & 50.24 & 2.13 & 0.50\\
     & RIS & 57.10 & 1.64 & 0.51 & 59.85 & 1.31 & 0.57 & 50.55 & 2.09 & 0.45 & 50.98 & 1.96 & 0.47 & 54.62 & 1.75 & 0.50\\
     & TO & 63.43 & 1.45 & 0.58 & 67.30 & 1.10 & 0.66 & 57.27 & 1.77 & 0.53 & 60.19 & 1.47 & 0.55 & 62.05 & 1.45 & 0.58\\
     & Paparazzo & \textbf{73.40} & \textbf{1.03} & \textbf{0.65} & \textbf{73.24} & \textbf{0.97} & \textbf{0.70} & \textbf{62.14} & \textbf{1.53} & \textbf{0.57} & \textbf{62.13} & \textbf{1.41} & \textbf{0.56} & \textbf{67.73} & \textbf{1.23} & \textbf{0.62}\\
    \midrule
    \multirow{4}{*}{\rotatebox{90}{SG}} & RW & 58.19 & 1.80 & 0.57 & 66.35 & 1.17 & 0.66 & 42.86 & 2.71 & 0.42 & 39.82 & 2.64 & 0.40 & 51.80 & 2.08 & 0.51\\
     & RIS & 51.49 & 2.09 & 0.50 & 57.23 & 1.40 & 0.57 & 44.36 & 2.71 & 0.43 & 41.23 & 2.34 & 0.41 & 48.58 & 2.14 & 0.48\\
     & TO & 65.42 & 1.31 & 0.63 & 67.27 & 1.09 & 0.66 & 57.66 & 1.65 & 0.56 & 52.82 & 1.60 & 0.52 & 60.79 & 1.41 & 0.59\\
     & Paparazzo & \textbf{73.27} & \textbf{1.04} & \textbf{0.69} & \textbf{72.97} & \textbf{0.97} & \textbf{0.71} & \textbf{64.07} & \textbf{1.41} & \textbf{0.61} & \textbf{61.82} & \textbf{1.38} & \textbf{0.59} & \textbf{68.03} & \textbf{1.20} & \textbf{0.65}\\
    \bottomrule
  \end{tabular}
  }
\end{table*}

\paragraph{Bouncing Ball~(BB) motion.} Although this case might be expected to be more challenging due to the object’s unpredictable motion when bouncing, reconstruction is generally easier. The object’s frequent bounces against walls allow it to be seen from multiple viewpoints, benefiting all methods. In this case, the Tracking-only~(TO) baseline achieves reasonably good results, reaching an average coverage of 75\%. However, Paparazzo consistently outperforms all baselines, achieving higher coverage than TO and surpassing 80\%.
\begin{figure*}[t]
    \centering
    \begin{subfigure}[t]{0.25\linewidth}
        \centering
        \includegraphics[width=\linewidth]{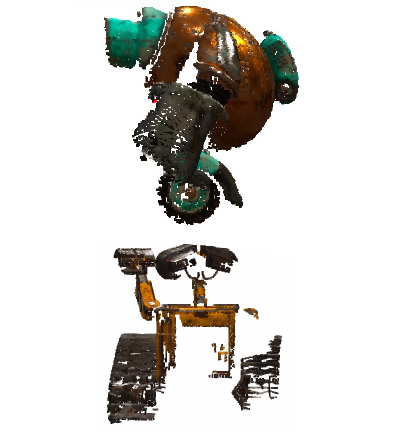}
        \caption{RW}
    \end{subfigure}
    \hfill
    \begin{subfigure}[t]{0.24\linewidth}
        \centering
        \includegraphics[width=\linewidth]{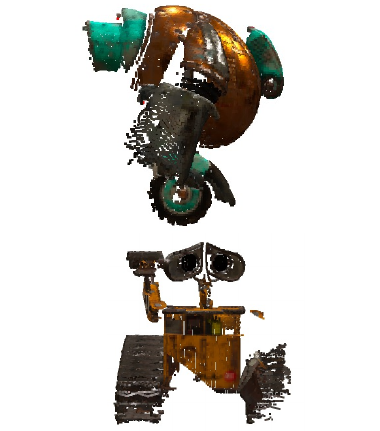}
        \caption{RIS}
    \end{subfigure}
    \hfill
    \begin{subfigure}[t]{0.25\linewidth}
        \centering
        \includegraphics[width=\linewidth]{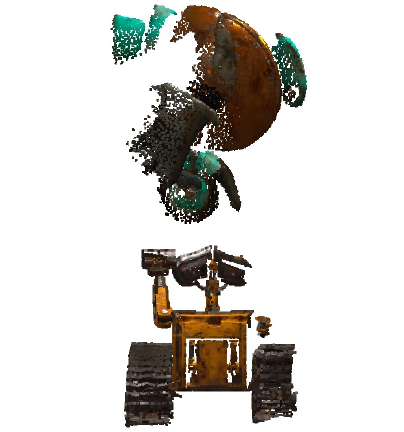}
        \caption{TO}
    \end{subfigure}
    \hfill
    \begin{subfigure}[t]{0.24\linewidth}
        \centering
        \includegraphics[width=\linewidth]{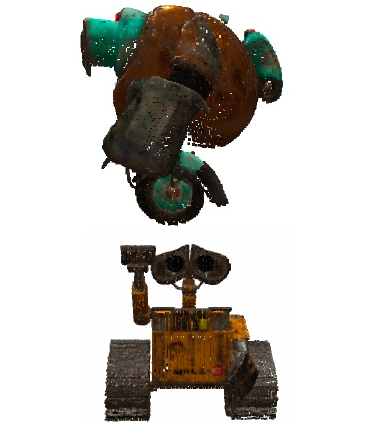}
        \caption{Paparazzo}
    \end{subfigure}

    \caption{\textbf{Visualization of the 3D reconstruction of Object 3 and Object 4 under Stop \& Go motion.} We compare the RW, RIS, and TO baselines against our Paparazzo method. Paparazzo produces significantly more complete and geometrically consistent reconstructions.}
    \label{fig:qualitative_res}
\end{figure*}
The advantage is even more evident when considering the AUC, where Paparazzo outperforms every baseline across almost all scenes, confirming its efficiency in dynamic object reconstruction. 

In contrast, the Random Walk~(RW) baseline performs poorly, as it completely disregards the object’s position. The Random Informative Selection~(RIS) baseline achieves slightly better results, but still underperforms compared to TO and Paparazzo, since it lacks synchronization with the object’s trajectory. Consequently, it frequently fails to reach feasible observation points in time, leading to incomplete reconstructions.

\paragraph{Curved Bouncing Ball~(CBB) motion.} This scenario reflects more real-world conditions, where the object follows curved trajectories and varies its speed, introducing unpredictability that makes trajectory estimation more challenging. As a result, Paparazzo’s performance drops significantly across all metrics. Notably, in the Ribera scene, Paparazzo achieves a coverage approximately 5\% lower than TO, likely due to the narrow environment, which limits the agent’s ability to anticipate the motion and reposition effectively before the next bounce, especially when the object moves faster. However, when averaged across all scenes, Paparazzo remains clearly superior for the CBB motion.

\paragraph{Forward~\&~Backward (FB) motion.} This condition is more challenging because the object does not rotate, requiring the agent to actively reason about the scene in order to reposition and observe it from new viewpoints. Consequently, performance metrics are generally lower. Nevertheless, Paparazzo achieves higher coverage than all baselines and a higher AUC across all scenes. TO performs worse due to its passive policy, which limits its ability to observe the object from diverse viewpoints. Finally, RW and RIS perform poorly due to their lack of object awareness and temporal anticipation.

\paragraph{Stop \& Go (SG) motion.} This represents the most challenging scenario, as the object intermittently pauses during motion, introducing unpredictable interruptions that hinder motion estimation. This motion pattern best reflects realistic conditions, where objects may temporarily stop or slow down. Consequently, all methods exhibit a general performance drop compared to the BB  motion.

For Paparazzo, difficulties arise when the agent has already planned to move toward an informative pose where the object is expected to appear, but the object stops earlier in a non-visible region. In such cases, the agent continues its motion without visual confirmation, making subsequent re-localization significantly more challenging. Nevertheless, Paparazzo still achieves substantially higher coverage and completeness than all baselines.

Paparazzo’s behavior allows it to effectively exploit stopping phases when the object remains visible, continuing the mapping and refining the reconstruction from different viewpoints  utilizing the available observation time. This advantage is reflected in the AUC, which increases by at least 10\% in almost all scenes compared to the best-performing baseline. In contrast, the Tracking-only (TO) mode remains idle, waiting for motion to resume, highlighting the limitations of a purely passive strategy, while RIS again performs poorly due to its lack of temporal synchronization.

\paragraph{}
Overall, these results highlight that Paparazzo’s adaptive alternation between Object Tracking Mode and Object Mapping Mode enables it to robustly handle diverse dynamic scenarios and motion complexities. By effectively balancing tracking accuracy with active exploration, Paparazzo consistently outperforms all baselines across motion types, leveraging motion variations and stop phases to achieve superior reconstruction performance.

To further assess the robustness and generality of the proposed method with respect to object variability, we report in Table~\ref{tab:motion_objects} the quantitative results obtained by Paparazzo and the baseline methods for each target object. 
This analysis is crucial to demonstrate that our approach generalizes well across diverse objects, especially since Paparazzo does not rely on object-specific training or fine-tuning. On average, Paparazzo improves coverage by nearly 10\% for Objects~1, 2, and 4. A smaller gain is observed for Object~3, where the improvement over the best-performing baseline (TO) is about 6.5\% on average. In particular, under the Curved Bouncing Ball~(CBB) motion, TO slightly outperforms Paparazzo in completeness while achieving comparable coverage of Object~3. This behavior is likely due to the object’s front-facing high-frequency texture, which biases the informative viewpoint selection toward similar orientations, reducing the effective diversity of the captured views. Nevertheless, Paparazzo still maintains the best overall average performance across all objects and metrics. The improvement is particularly significant in the Stop \& Go~(SG) motion, where Paparazzo achieves more than 10\% higher coverage and AUC than the best-performing baseline. This result highlights its ability to dynamically balance exploration and reconstruction under less predictable or partially observable object trajectories (see~\cref{fig:qualitative_res}).

All these findings confirm that Paparazzo’s policy, alternating between object tracking and active mapping, generalizes effectively across both shape and appearance variations. Its design allows robust handling of heterogeneous objects and motion behaviors without any task-specific training, demonstrating strong potential for deployment in real-world dynamic reconstruction scenarios.

\section{Conclusion}
\label{sec:conclusion}


In this work, we introduced the new task of active mapping of a rigid moving object, a setting that departs from the long-standing assumption of static scenes in exploration and reconstruction. 

We proposed Paparazzo, a learning-free framework that integrates motion prediction, information-driven viewpoint selection, and behavioral adaptation to the target object’s dynamics. Our experiments show that Paparazzo substantially surpasses existing baselines, demonstrating that explicitly reasoning about where to look, when to look, and how to adapt the agent’s behavior to the object’s motion is crucial for efficient and accurate reconstruction of moving objects. By continuously balancing viewpoint informativeness, motion feasibility, and field-of-view maintenance, Paparazzo enables a form of dynamic scene understanding that was previously unexplored.

We believe our work establishes an important foundation: active viewpoint planning coupled with motion-aware behavior is key for bringing 3D reconstruction beyond static scenes and toward real dynamic environments. 

\section*{Acknowledgment}
This work was in part supported by the European Union (ERC Advanced Grant explorer Funding ID \#101097259)

{
    \small
    \bibliographystyle{ieeenat_fullname}
    \bibliography{main}
}

\clearpage
\setcounter{page}{11}
\maketitlesupplementary
\appendix

\section{Additional Details on Paparazzo}
\label{sec:add_details_method}

This section provides additional technical details on the Paparazzo framework.
First, we present the complete formulation of the Extended Kalman Filter (EKF) used for motion prediction, together with the process/measurement noise parameters and the confidence criteria that drive the switch between Object Tracking Mode and Object Mapping Mode (\cref{sec:ekf_formulation}).
Next, we describe the procedure used to accumulate the object point cloud over time (\cref{sec:object_pcd}).
Finally, we detail how the dynamic object Gaussian model $\mathcal{G}_{O}$ is incrementally updated using the current observation and selected past keyframes (\cref{sec:gauss_update_dyn}).

\subsection{EKF Formulation on \texorpdfstring{$SE(3)$}{SE(3)}}
\label{sec:ekf_formulation}

The EKF used in Paparazzo estimates the object pose in the world reference frame and its body-frame twist:
\begin{equation}
(\,T_k,\; \omega_k\,), 
\qquad
T_k \in SE(3),\;\; \omega_k \in \mathbb{R}^6 \> .
\end{equation}

Since \(SE(3)\) is a nonlinear manifold, the EKF operates on a minimal error state expressed in its tangent space:
\begin{equation}
\delta x_k =
\begin{bmatrix}
\delta\xi_k \\[2pt]
\delta\omega_k
\end{bmatrix}
\in \mathbb{R}^{12},
\qquad
\delta\xi_k \in \mathfrak{se}(3),\;
\delta\omega_k \in \mathbb{R}^6 \> .
\end{equation}
Here, $\delta\xi_k$ represents a 6-DoF perturbation of the pose expressed in the Lie algebra $\mathfrak{se}(3)$, while $\delta\omega_k$ encodes the deviation in the object twist.
With this representation, the EKF proceeds through the standard prediction and update phases using Lie-group consistent linearizations.

\paragraph{Prediction.}
Under the constant-twist motion model adopted in Paparazzo, the pose evolves via the exponential map:
\begin{equation}
T_{k|k-1} = T_{k-1}\,\exp\!\bigl(\omega_{k-1}\,\Delta t\bigr) \> .
\end{equation}
The covariance follows the linearized dynamics:
\begin{equation}
P_{k|k-1} = F_k P_{k-1} F_k^\top + Q \> ,
\end{equation}
where \(F_k\) is the Jacobian of the motion model and \(Q\) is the block-diagonal
process noise matrix defined in \cref{sec:ekf_details}.
Here, $P_{k-1}$ denotes the posterior covariance at time step $k\!-\!1$ and $P_{k|k-1}$ the predicted covariance at time step $k$ before incorporating the new measurement.
\paragraph{Update.}
Whenever the object is visible, Paparazzo provides an absolute pose estimate \(T_{O_k}^{W,\mathrm{meas}}\in SE(3)\), obtained from the pose initialization and the ICP refinement (as described in \cref{sec:track_mode} and detailed in \cref{sec:object_pcd}).  
The innovation is computed on the Lie algebra:
\begin{equation}
y_k = \log(T_{\text{err}}) \in \mathfrak{se}(3) \> .
\end{equation}
where $T_{\text{err}} = (T_{k|k-1})^{-1} T^{W,\text{meas}}_{O_{k}}$.

Since the measurement constrains only the object pose, the Jacobian is
\begin{equation}
H = \begin{bmatrix} I_6 & 0 \end{bmatrix} \> ,
\end{equation}
and the innovation covariance and Kalman gain are respectively
\begin{equation}
    S_k = H P_{k|k-1} H^\top + R,
    \qquad
    K_k = P_{k|k-1} H^\top S_k^{-1} \> .
\end{equation}

The error state and covariance update follow the standard EKF equations:
\begin{equation}
\delta x_k = K_k\, y_k,
\qquad
P_k = ({I}-K_k H)\,P_{k|k-1} \> ,
\end{equation}
where $P_k \equiv P_{k|k}$ denotes the posterior covariance after the update at time step $k$. 
The nominal state is then corrected by retracting the pose increment on \(SE(3)\):
\begin{equation}
T_k = T_{k|k-1}\,\exp(\delta\xi_k)\>,
\qquad
\omega_k = \omega_{k|k-1} + \delta\omega_k\>.
\end{equation}

This formulation explicitly shows how the process noise $Q$, measurement noise $R$, and innovation $y_k$ contribute to the filtering process.


\subsubsection{EKF Parameters and Confidence Criteria}
\label{sec:ekf_details}

The process noise $Q$ is block–diagonal, separating uncertainty in pose and twist:
\begin{equation}
{Q} =
\begin{bmatrix}
{Q}_t & 0 \\
0 & {Q}_\omega
\end{bmatrix},
\qquad
{Q}_t = 10^{-4}{I}_6 \>,
\end{equation}
\[
{Q}_\omega =
\mathrm{diag}\!\big(10^{-4},10^{-4},10^{-4},\,
                   5\!\times\!10^{-4},5\!\times\!10^{-4},5\!\times\!10^{-4}\big)\>.
\]

These values are deliberately small: the object moves smoothly from frame to frame and receives frequent and accurate pose corrections from ICP.
Slightly larger rotational entries in $Q_\omega$ keep the filter responsive to small angular variations while maintaining stability.
All parameters were tuned empirically to balance prediction smoothness with robustness against occasional ICP imperfections (e.g., partial views or local registration jitter).
The measurement noise covariance is set to $R = 10^{-3} I_6$, modeling the uncertainty of the absolute 6D pose provided by ICP.
Since the filter starts with no prior knowledge of the object state, the initial covariance is set to $P_0 = I_{12}$, representing an uninformative prior over pose and twist.

Given these noise models, the key question becomes determining when the EKF is sufficiently reliable to switch from Object Tracking Mode to Object Mapping Mode.
To this end, Paparazzo monitors two complementary quantities derived from the EKF state:
(i) the uncertainty $U_k$, and
(ii) the innovation consistency metric $\mathrm{NIS}_k$.
These values are compared against their respective thresholds, $\tau_u = 0.1$ and $\tau_n = 0.5$, as detailed in \cref{sec:ekf}. In particular, the choice of the NIS threshold is guided by empirical observations.
Although a 6D pose measurement would theoretically call for a $\chi^2(6)$ statistical test, in practice the innovations are orders of magnitude smaller. This is a direct consequence of three conditions in our setting: the object exhibits smooth frame-to-frame motion, the process noise is intentionally low to preserve rigid-motion coherence, and the ICP refinement provides accurate local pose corrections.
Therefore, classical chi-squared thresholds (e.g., 12.59 at the 95th percentile) are too large to be meaningful. Instead, we adopt a data-driven threshold: during steady motion the NIS remains consistently below $0.1{-}0.2$, whereas genuine motion changes or ICP inconsistencies yield significantly larger values.
Based on this empirical separation, we set $\tau_n = 0.5$. 
Thus, values above this threshold indicate that the observed motion is no longer compatible with the current state estimate.
However, using these thresholds at a single time step is not sufficient, rather, they serve as indicators for assessing long-term EKF stability.

\paragraph{Confidence Condition.}
The EKF is considered confident when both the uncertainty and innovation remain below their thresholds for \(N_s = 4\) consecutive frames:
\begin{equation}
  U_k < \tau_u,\qquad \mathrm{NIS}_k < \tau_n\,.  
\end{equation}
This temporal stability requirement prevents spurious mode switches caused by short-term ICP mismatches or partial occlusions. Only after \(N_s\) stable frames the system transitions from the Object Tracking Mode to the Object Mapping Mode. Conversely, if innovation quantity exceeds its threshold for \(N_s\) consecutive frames, the system reliably detects a genuine motion change and switches back to the reactive Object Tracking Mode.

\Cref{tab:ekf_params} summarizes all EKF parameters.

\begin{table}[h]
\centering
\caption{EKF parameters used in all experiments.}
\resizebox{0.47\textwidth}{!}{%
\begin{tabular}{l c}
\toprule
Parameter & Value \\
\midrule
$\tau_u$ (uncertainty threshold) & 0.1 \\
$\tau_n$ (NIS threshold) & 0.5 \\
$N_s$ (stability window) & 4 \\
${Q}_t$ & $10^{-4}{I}_6$ \\
${Q}_\omega$ & diag$(10^{-4},10^{-4},10^{-4},5\!\cdot\!10^{-4},5\!\cdot\!10^{-4},5\!\cdot\!10^{-4})$ \\
${R}$ & $10^{-3}{I}_6$ \\
$P_{0}$ & $I_{12}$\\
\bottomrule
\end{tabular}
}
\label{tab:ekf_params}
\end{table}

\subsection{Object Point Cloud Accumulation}
\label{sec:object_pcd}

The final object point cloud $\mathcal{P}$ is expressed in the reference frame of the object at its first detection time $t_d$. Whenever the object is visible and the mask $\mathcal{M}_k$ is available, we extract the corresponding object point cloud $\mathcal{P}^{C_k}_{O_k}$ in the current camera frame $C_k$. To bring this point cloud into the object reference frame $O_{t_{d}}$, we compute the following initial alignment:
\begin{equation}
\label{eq:align}
\hat{\mathcal{P}}^{O_{{t_{d}}}}_{O_k}
= 
T_{W}^{O_{t_d}}\,
T_{C_k}^{W}\,
\mathcal{P}^{C_k}_{O_k}\>,
\end{equation}
where $T_{W}^{O_{t_d}}$, is the inverse of the rototranslation computed during the initialization phase (\cref{sec:initialization}).  
However, the point cloud $\hat{\mathcal{P}}^{O_{{t_{d}}}}_{O_k}$ does not yet account for the motion that occurred between the detection time $t_d$ and the current step $k$. As a result, it remains misaligned with the previously accumulated 3D model $\mathcal{P}$. To recover this missing relative motion and correctly integrate the new observations, we apply the alignment procedure described in \cref{sec:track_mode}. We first compute a coarse but globally consistent registration using KISS-Matcher, which is robust to outliers and large inter-frame displacements, estimating an initial transformation $\hat{T}_{\text{align},k}$. This estimate is then refined with Colored ICP to obtain the final transformation $T_{\text{align},k}$, that is applied to $\hat{\mathcal{P}}^{O_{{t_{d}}}}_{O_k}$ before merging it into the current object model:
\begin{equation}
\mathcal{P} \leftarrow \mathcal{P} \cup (T_{\text{align},k}\,\hat{\mathcal{P}}^{O_{{t_{d}}}}_{O_k}).    
\end{equation}
The same alignment also provides the absolute 6D pose measurement used by the EKF
\begin{equation}
T^{W,\text{meas}}_{O_{k}}=T^{W}_{O_{t_{d}}}\,T_{\text{align},k}\,.
\end{equation}

\subsection{Gaussian Optimization for the Dynamic Object}
\label{sec:gauss_update_dyn}

Whenever the object is detected and its mask $\mathcal{M}_k$ is available, we update the dynamic Gaussian model $\mathcal{G}_O$. 
At time $k$, all Gaussians are expressed in the current object reference frame $O_k$, which evolves over time according to the relative motions $T_{O_k}^{O_{t_{d}}}$ obtained during the alignment stage.
Before optimizing the model, Paparazzo performs densification, which requires determining which regions of the object surface are actually visible at time $k$ and already represented in $\mathcal{G}_O$.
To do this, all Gaussian centers $g^{O_k}\in\mathbb{R}^{3}$ are first transformed into the current camera frame $C_k$ using the latest estimated object pose:
\begin{equation}
{g}^{C_k} 
= 
T^{C_k}_{W}\,
T^{W,\text{meas}}_{O_k}\,{g}^{O_k}\,.
\end{equation}
The projection enables silhouette rendering:
pixels in the mask $\mathcal{M}_k$ that are not covered by any projected Gaussian indicate unexplained regions of the object surface, where new Gaussians must be inserted. However, mask-based visibility alone is insufficient. Gaussian centers corresponding to the back side of the object may still fall inside $\mathcal{M}_k$ when projected, even though they are not physically visible in the current view.
To ensure geometric correctness, Paparazzo applies a depth-consistency filter. For each projected Gaussian at pixel $u$, let $ z_{g_k}(u)$ be its depth and $z_k(u)$ the depth observed in the current frame.
A Gaussian is considered visible and already accounted for in $\mathcal{G}_O$ only if:
\begin{equation}
    \bigl|z_{g_k}(u) - z_k(u)\bigr| \le \tau_d\,,
\qquad\tau_d = 0.02\text{ m}\,.
\end{equation}
Here, $\tau_d$ defines a small but reasonably flexible depth tolerance, which can be relaxed if needed to accommodate larger discrepancies caused by sensor noise and slight registration errors.
This constraint ensures that only the physically observable surface contributes to determining which parts of the object require densification.

\paragraph{Keyframe Selection and Optimization}
For the gaussian model optimization, Paparazzo leverages past keyframes that still observe the 
same surface region visible at time~$k$.
Each past keyframe $i<k$ stores an object point cloud $\mathcal{P}^{C_i}_{O_i}$ extracted at time~$i$.  
During the accumulation procedure, as described in \cref{sec:object_pcd},
this point cloud is first expressed in the object reference frame $O_{t_d}$ (using
\cref{eq:align}), and then registered via the estimated alignment transformation
$T_{\text{align},i}$:
\begin{equation}
    \hat{\mathcal{P}}_{i}
=
T_{\text{align},i}\,\hat{\mathcal{P}}^{O_{t_d}}_{O_i}\,.
\end{equation}

\begin{figure}[t]
    \centering
    \includegraphics[width=1.0\linewidth]{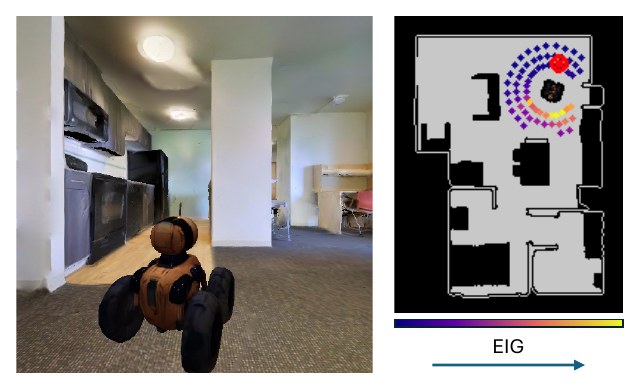}
    \caption{\textbf{Generation of candidate viewpoints for Object Mapping Mode.}
    When the EKF becomes confident, Paparazzo switches from tracking to mapping and evaluates a set of candidate viewpoints $\mathcal{V}$ distributed around the object.
    The expected information gain (EIG) of each pose, computed using the FisherRF criterion, is visualized here with a color gradient: darker tones correspond to low informativeness, while brighter tones highlight more informative poses for reconstructing the object.}
    \label{fig:pose_gen}
\end{figure}

To determine whether keyframe $i$ observes the same part of the object currently visible at time $k$, we transform $\hat{\mathcal{P}}_i$ into the current camera frame $C_k$:
\begin{equation}
\mathcal{P}^{C_k}_{O_i}=T^{C_k}_{W}\, T^{W}_{O_{t_{d}}}\,
\hat{\mathcal{P}}_{i}\,.
\end{equation}

Then, we reproject $\mathcal{P}^{C_k}_{O_i}$ into the current image plane, obtaining a set of pixel locations $\Omega_{k,i}$. We retain only the pixels that fall inside the current mask:
\begin{equation}
\Omega_{k,i} \leftarrow \Omega_{k,i} \cap \mathcal{M}_{k}\,.    
\end{equation}

Even in this case, mask agreement alone is insufficient, since points from the back side of the 
object may still project inside $\mathcal{M}_k$.  
Therefore, we apply the same depth-consistency filter used during densification:
a reprojected point at pixel $u$ is kept only if its depth is consistent with the 
current observation,
\begin{equation}
\bigl|\,z_{i \rightarrow k}(u) - z_k(u)\,\bigr|
\;\le\;
\tau_d,
\end{equation}
where $z_{i \rightarrow k}(u)$ is the depth of the reprojected 3D point in the camera frame $C_{k}$ from keyframe $i$.

Let $\Omega_{k,i}'$ denote the set of points passing both mask and depth checks.
We compute the visibility overlap
\begin{equation}
\eta_i =
\frac{|\Omega_{k,i}'|}{|\mathcal{M}_k|} \> .
\end{equation}

Only keyframes with $\eta_i \ge 0.5$ are retained, and at most the top
20 highest-overlap ones are selected for the gaussian optimization.  
Because the object is rigid, this 3D-aware visibility test ensures that only
keyframes observing the same surface region as the current view
meaningfully contribute to the update.

We then jointly optimize the Gaussians associated with the selected past keyframes
and the current one, using RGB and SSIM losses
\cite{keetha2024splatam}.  
This produces a spatially localized, temporally consistent, and computationally
efficient refinement of the dynamic object model.

\section{Additional Details on Experimental Results}

\begin{figure}[t]
    \centering
    \includegraphics[width=1.0\linewidth]{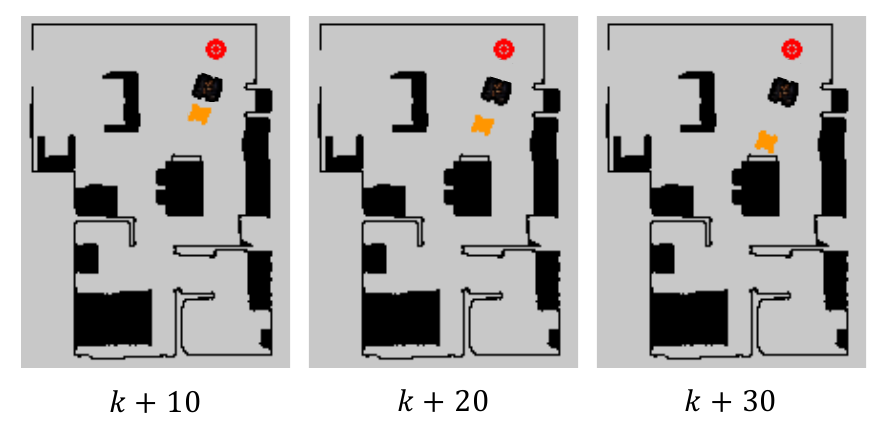}
    \caption{\textbf{EKF-based prediction of future object poses.}
    The EKF predicts the object pose, denoted in orange, over the next $N_h$ steps.
    The figure shows three examples at steps $k{+}10$, $k{+}20$, and $k{+}30$.
    These predicted poses are then used to propagate all candidate viewpoints $\mathcal{V}$ into the future, producing a set of $|\mathcal{V}|\times N_h$ future-aligned viewpoints that are subsequently evaluated by the final cost function in \cref{eq:cost_function_fin}.}
    \label{fig:future_pred}
\end{figure}


To complement the evaluation presented in \cref{sec:exp_and_res}, this section provides additional technical details, further qualitative comparisons, and examples of representative agent and object trajectories.  
These materials are intended to give a deeper understanding of the behavior of Paparazzo and the baseline methods across different scenarios.

\paragraph{Experimental setup and runtime details.}
All experiments use a $512 \times 512$ RGB-D camera with a $90^\circ$ field of view, consistent with prior active mapping work. Paparazzo runs online at approximately 8 FPS while using at most 4\,GB of GPU memory. The EKF update requires at most 5\,ms per step, while joint 3DGS refinement and FisherRF evaluation require about 0.1\,s per step. The A* planner is not executed at every step; it is triggered intermittently, e.g., when switching from tracking to mapping mode or when a target viewpoint has been reached, and requires at most 0.5\,s per invocation. All experiments were conducted on a single NVIDIA GeForce RTX 4090 GPU.

\paragraph{Paparazzo details.}
During Object Tracking Mode, the agent keeps the object within the camera's field of view by performing small corrective rotations of $\pm 10^\circ$, as described in \cref{sec:exp_and_res}.  
The agent also maintains a distance of 1.5–2.5\,m from the object.  
This emulates a realistic “follow-and-observe” behavior and allows reliable short-term tracking even in compact environments.

To construct the viewpoint candidates $\mathcal{V}$ for Object Mapping Mode, candidate camera poses are sampled around the object using a foveated, object-centered distribution.
Camera centers lie on three concentric circular rings on a common horizontal plane, with radii between $1.2$ and $1.8\,\text{m}$. Samples are placed every $12^\circ$ in azimuth, and all camera orientations point toward the object (see \cref{fig:pose_gen}). This design provides dense and uniform surface coverage while restricting viewpoints to geometrically feasible and navigable regions of the scene. Each candidate viewpoint is then evaluated using the FisherRF criterion, which quantifies its expected information gain (EIG).

Because the object is dynamic, Paparazzo must reason not only about the informativeness of a viewpoint at the current time but also about how informative it will remain as the object moves.
To do this, Paparazzo propagates all candidate viewpoints $\mathcal{V}$ across the next $N_h=60$ predicted object poses obtained from the confident EKF (see \cref{fig:future_pred}).
This results in a set of $|\mathcal{V}| \times N_h$ future-consistent candidate viewpoints, each synchronized with the motion of the object and equipped with its predicted EIG.

Finally, Paparazzo selects the target viewpoint using the cost function defined in \cref{eq:cost_function_fin}, which jointly accounts for information gain, reachability, and temporal synchronization between the agent trajectory and the object’s predicted motion.
In all experiments, the weights in \cref{eq:cost_function_fin} are set to \(w_\eig = 0.8\) and \(w_\sync = 1.2\), empirically determined on a validation set. 

For each run, all methods start from the same object initialization. The object is randomly placed in front of the agent at a distance in the range [1.0, 2.5] m, with a lateral offset in [-0.5, 0.5] m relative to the agent, and a random orientation in [0, 360) around the z-axis of its own reference frame.

\subsection{Additional Results}
\begin{figure}[!t]
    \centering
    \includegraphics[width=\linewidth]{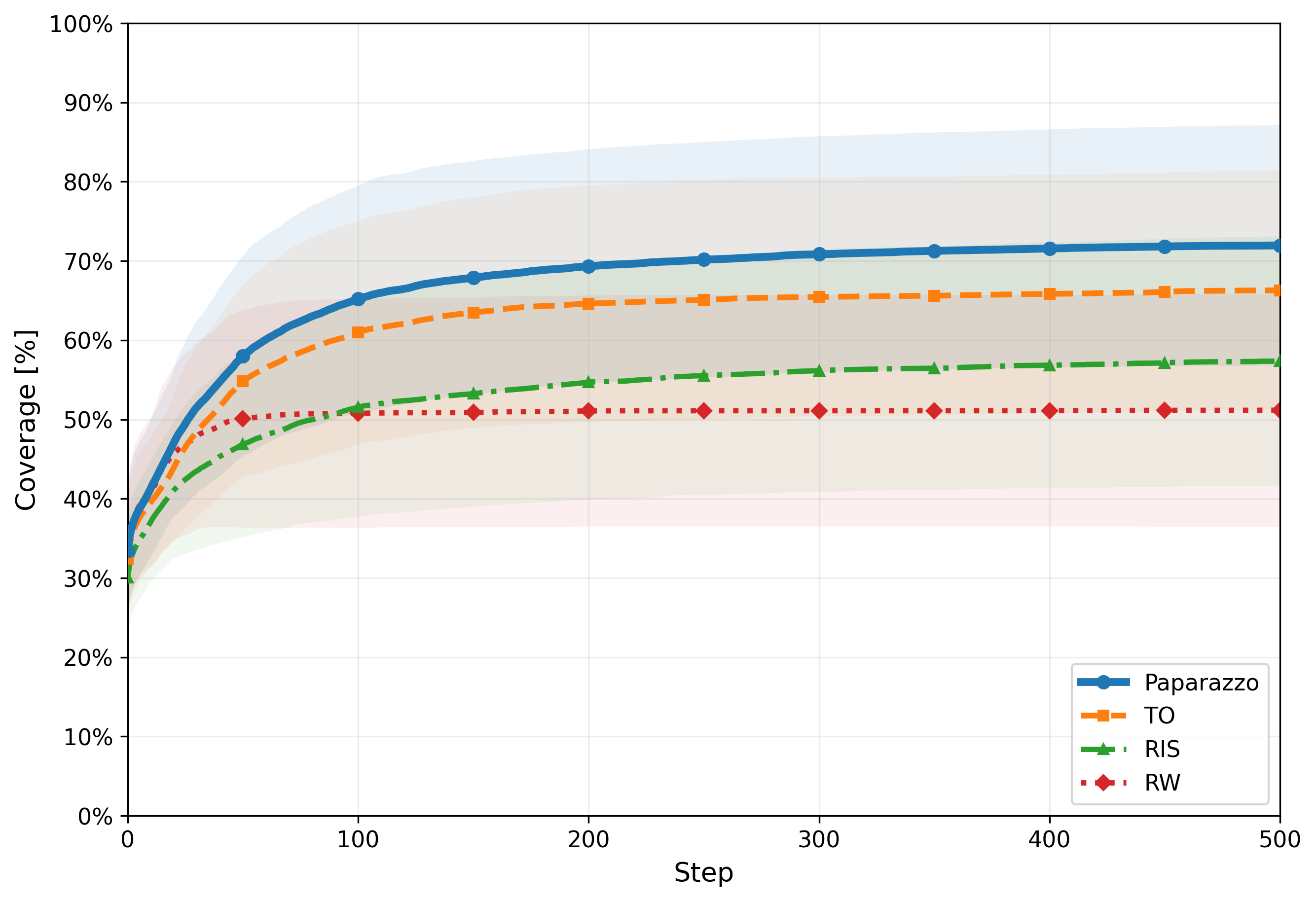}
    \caption{\textbf{Coverage over exploration steps.} Results are averaged across all scenes, motion patterns, and objects. Paparazzo consistently achieves higher coverage throughout the entire step budget. Shaded areas indicate the standard deviation across runs.}
    \label{fig:coverage_vs_step_all}
\end{figure}

To further visualize the advantages of Paparazzo over the RW, RIS, and TO baselines, \cref{fig:coverage_vs_step_all} reports the coverage as a function of the exploration steps, averaged across all scenes, motion patterns, and objects. This plot clearly illustrates the benefit of achieving a higher AUC: a method with a larger AUC reconstructs the object more efficiently throughout exploration, rather than only reaching a higher final coverage. As shown in the figure, Paparazzo consistently maintains higher coverage over the entire step budget, confirming its ability to acquire informative viewpoints earlier and more effectively than the baselines.

\begin{figure*}[t]
    \centering
    \begin{subfigure}[t]{0.24\linewidth}
        \centering
        \includegraphics[width=\linewidth]{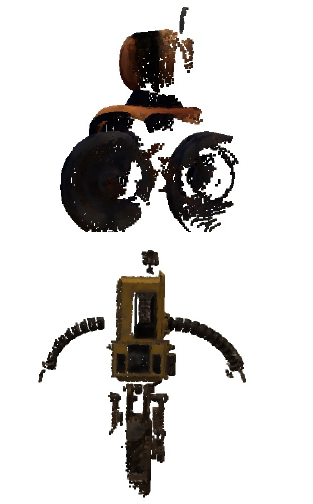}
        \caption{RW}
    \end{subfigure}
    \hfill
    \begin{subfigure}[t]{0.24\linewidth}
        \centering
        \includegraphics[width=\linewidth]{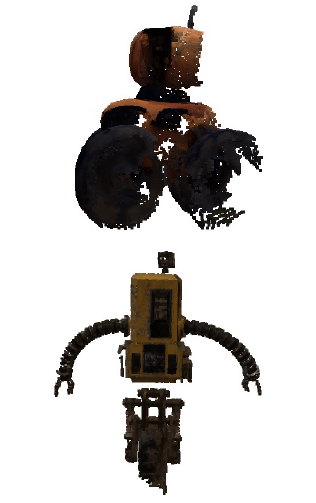}
        \caption{RIS}
    \end{subfigure}
    \hfill
    \begin{subfigure}[t]{0.225\linewidth}
        \centering
        \includegraphics[width=\linewidth]{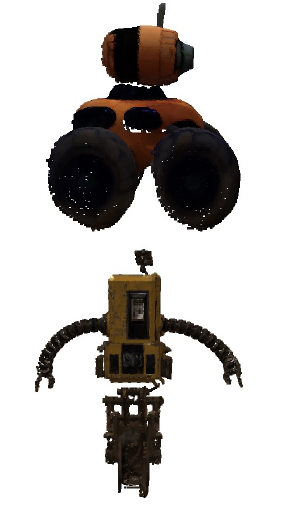}
        \caption{TO}
    \end{subfigure}
    \hfill
    \begin{subfigure}[t]{0.24\linewidth}
        \centering
        \includegraphics[width=\linewidth]{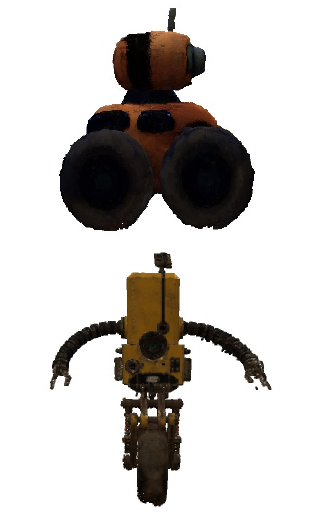}
        \caption{Paparazzo}
    \end{subfigure}

    \caption{\textbf{Qualitative 3D reconstructions of Object~1 and Object~2 under Stop \& Go motion.} Paparazzo consistently produces reconstructions that are significantly more complete, coherent, and stable across both objects. In contrast, the baselines, especially on Object~2, fail to recover the entire frontal surface, leaving substantial portions missing.}
    \label{fig:qualitative_res_sup_mat}
\end{figure*}

This quantitative advantage is also reflected in the final reconstruction quality. To complement the coverage-over-time analysis, additional qualitative reconstructions are shown in \cref{fig:qualitative_res_sup_mat}.

Across all tested objects, Paparazzo produces reconstructions that are both more complete and geometrically more consistent, particularly under challenging motion patterns. This is especially evident in the Stop \& Go setting, where baseline methods frequently fail to recover large portions of the object due to motion discontinuities or prolonged static intervals.

In the examples of Object1 and Object2 reported in \cref{fig:qualitative_res_sup_mat}, Paparazzo achieves substantially higher completeness, reaching 81.23\% and 74.45\%, respectively. In contrast, all baseline methods exhibit severe missing regions. Notably, for Object~2 the RW, RIS, and TO baselines consistently fail to reconstruct the entire frontal surface of the object, demonstrating their difficulty in handling partial observations and non-smooth trajectories.
Paparazzo, instead, successfully recovers these surfaces thanks to its dynamic viewpoint selection and continuous information-driven mapping strategy, which remain effective even when the object stops or follows irregular motions.

The coverage values obtained in this experiment for all methods are summarized in \cref{tab:comp_supmat}.

\begin{table}[h]
    \centering
    \setlength{\tabcolsep}{10pt}
    \renewcommand{\arraystretch}{1.2}
    \caption{Reconstruction coverage (\%) for Object~1 and Object~2 under the \textit{Stop \& Go} motion pattern. Paparazzo significantly outperforms all baselines.}
    \begin{tabular}{lcc}
        \toprule
        \textbf{Method} & \textbf{Object 1} & \textbf{Object 2} \\
        \midrule
        RW  & 43.24 & 39.42 \\
        RIS & 51.12 & 49.23 \\
        TO  & 68.25 & 58.24 \\
        Paparazzo & \textbf{81.23} & \textbf{74.45} \\
        \bottomrule
    \end{tabular}
    \label{tab:comp_supmat}
\end{table}

\begin{figure*}
    \centering \captionsetup{type=figure} 
    \includegraphics[width=0.93\textwidth]{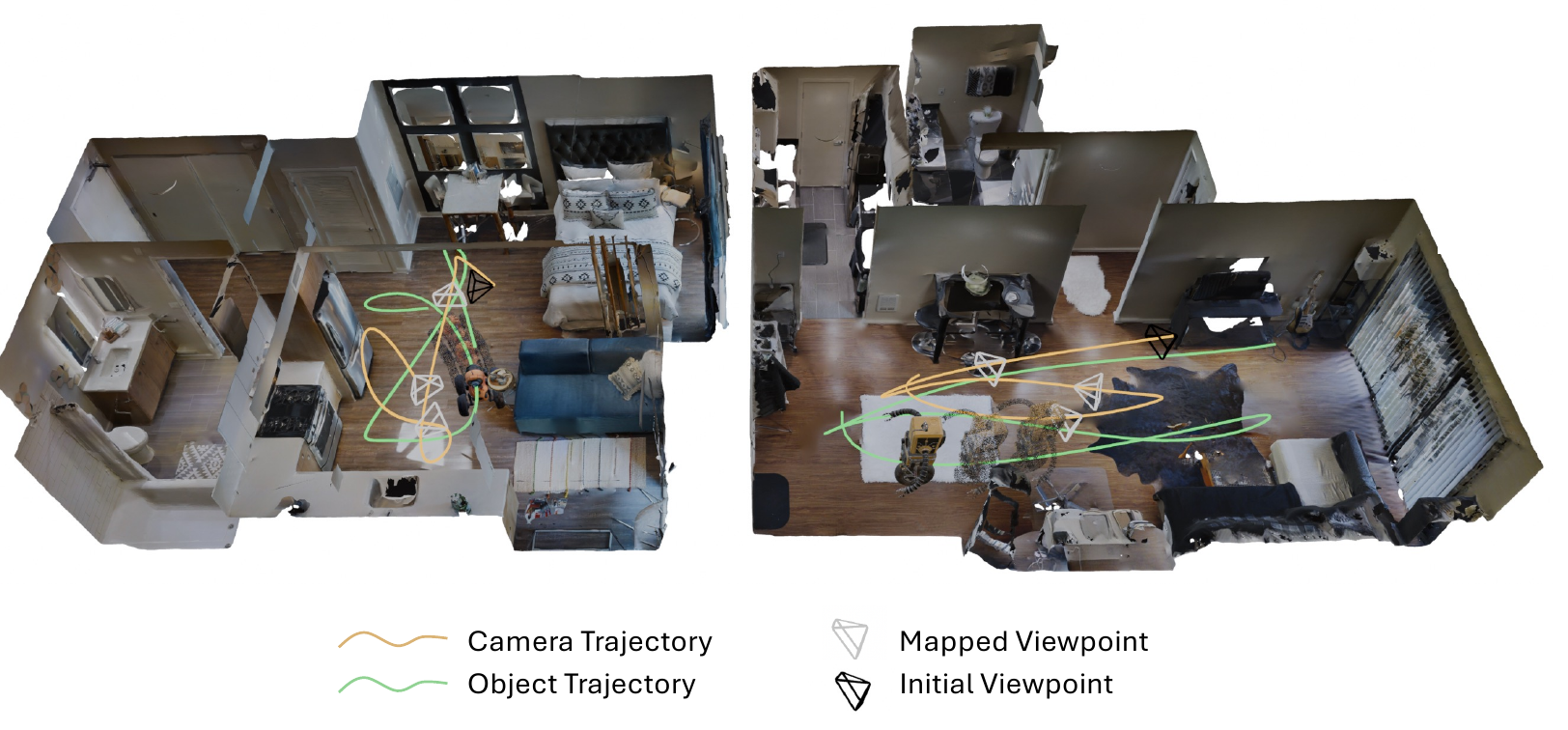}\caption{Benchmark examples of active mapping of moving objects. In each scenario, the agent plans camera viewpoints around a moving target while compensating for its motion to acquire informative observations for reconstruction.}
    \label{fig:demo}
\end{figure*}

\begin{figure*}[t]
    \centering
    \includegraphics[width=1.0\linewidth]{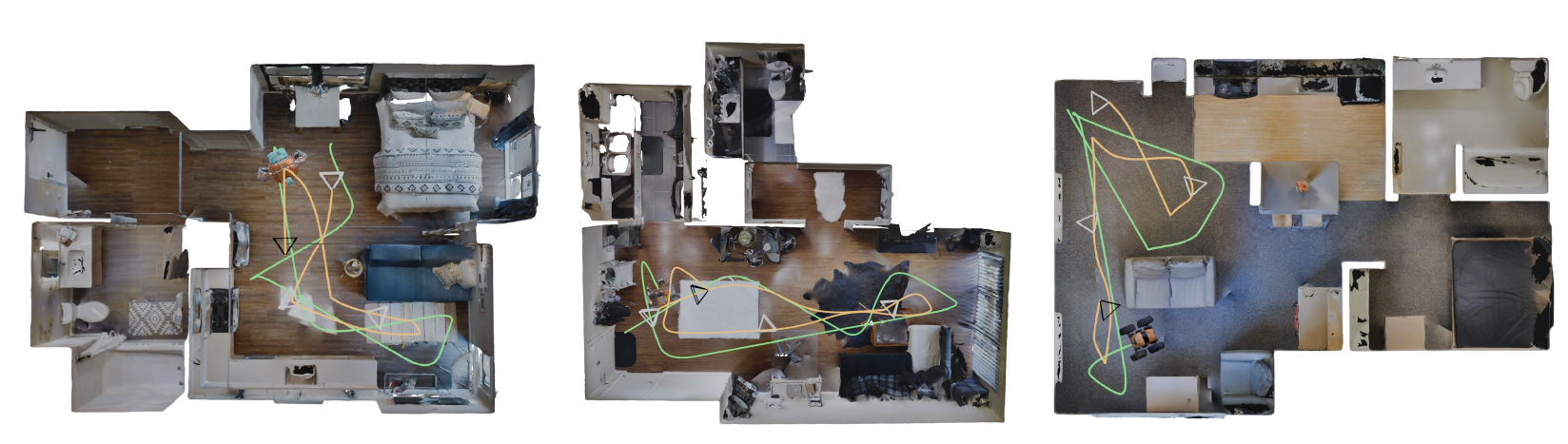}
    \caption{\textbf{Examples of object trajectories}. The moving target is performing the Bouncing Ball motion in the Denmark, Greigsville, and Ribera scenes (left to right). The agent executes the Paparazzo framework while continuously adapting its motion to track and map the moving object.}
    \label{fig:bb_traj}
\end{figure*}

\begin{figure*}[t]
    \centering
    
    \begin{subfigure}[t]{0.48\linewidth}
        \centering
        \includegraphics[width=\linewidth]{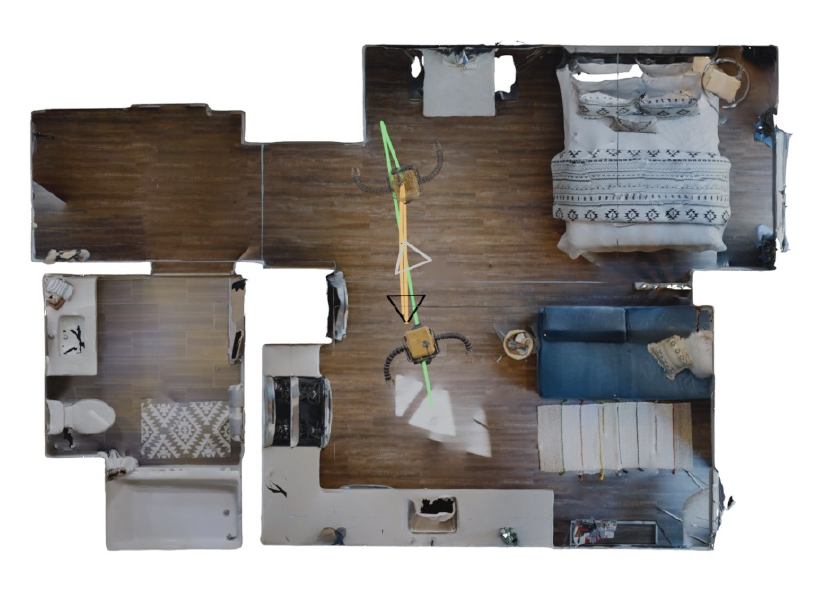}
        \caption{TO}
        \label{fig:traj_track}
    \end{subfigure}
    \hfill
    \begin{subfigure}[t]{0.48\linewidth}
        \centering
        \includegraphics[width=\linewidth]{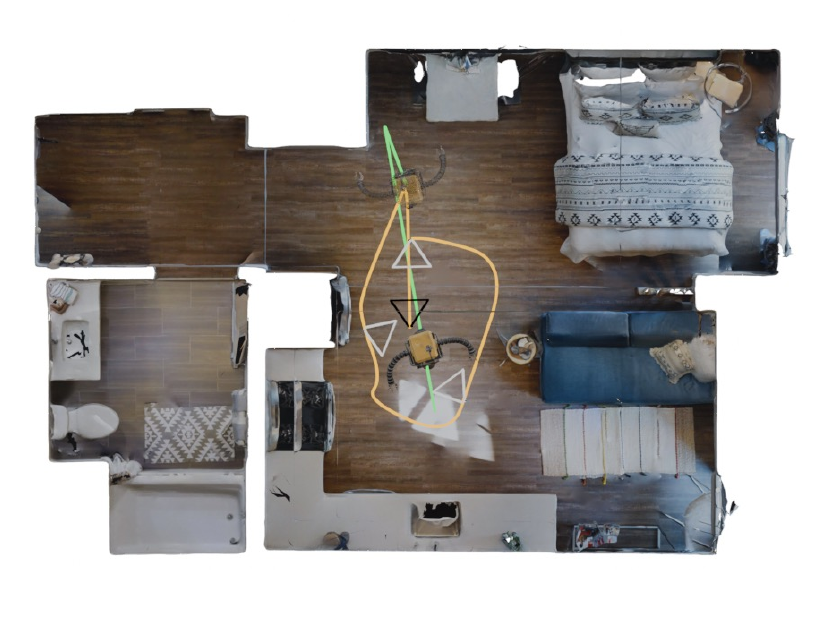}
        \caption{Paparazzo}
        \label{fig:traj_papa}
    \end{subfigure}

    \caption{\textbf{Illustration of Stop \& Go motion.} 
    Both panels show the same Stop \& Go trajectory executed by the moving object. Across 300 steps, the object pauses once. (a) TO: the agent remains passive during the stop phase, losing valuable time and collecting no new viewpoints, which prevents further progress in the reconstruction. (b) Paparazzo: the agent continues to actively reposition and capture informative views even while the object is stationary, as visible from the additional exploratory camera frustums (in gray). This allows Paparazzo to maintain reconstruction progress during motion interruptions, unlike the TO baseline.}
    \label{fig:comparison}
\end{figure*}

\subsection{Additional Analysis}
\label{sec:add_analysis}

To provide a deeper understanding of Paparazzo’s behavior during reconstruction,  \cref{fig:demo} illustrates two example scenarios from our benchmark.
Additionally, in \cref{fig:bb_traj} we show three examples where the object follows a Bouncing Ball (BB) motion across three distinct scenes---Denmark, Greigsville, and Ribera.
The black frustum marks the initial camera pose, after which the agent alternates between Object Tracking Mode and Object Mapping Mode, continuously adapting its trajectory to maximize reconstruction quality.
These examples highlight the robustness of our approach: Paparazzo consistently produces purposeful and informative motion across different environments and object types, without requiring any object-specific tuning or prior scene knowledge.

For completeness, \cref{fig:comparison} presents an example of a Stop \& Go motion sequence in the Denmark scene, visualized over the first 300 steps for clarity.
The blurred instance of the object marks its initial position, whereas the sharper instance indicates the location where it remained stationary for S=100 steps.

This qualitative comparison highlights a key limitation of the Tracking-Only (TO) baseline (\cref{fig:traj_track}): although TO operates effectively when the object is continuously moving, it fails when the object stops. In these situations, TO simply remains static, illustrated by the grey camera frustum, waiting passively for motion to resume.

In contrast, Paparazzo (\cref{fig:traj_papa}) continues reconstructing autonomously even during long standstills. It keeps evaluating informative viewpoints, navigating toward them, and acquiring new observations. As a result, Paparazzo maintains reconstruction progress and achieves a more complete and temporally consistent model of the object, despite the absence of motion.



    
    

\end{document}